\documentclass[journal]{IEEEtran}
\ifCLASSINFOpdf
   \usepackage[pdftex]{graphicx}
\else
   \usepackage[dvips]{graphicx}
   \graphicspath{{../eps/}}
\fi
\usepackage{soul}
\usepackage{threeparttable}
\usepackage{amsfonts}
\usepackage{graphicx}
\usepackage{subfigure}
\usepackage[justification=centering]{caption}
\ifCLASSOPTIONcompsoc
  \usepackage[caption=false,font=normalsize,labelfont=sf,textfont=sf]{subfig}
\else
  \usepackage[caption=false,font=footnotesize]{subfig}
\fi
\begin{document}
%
\title{Generic Model-Agnostic Convolutional Neural Network for Single Image Dehazing}
%
%
%

\author{Zheng Liu,
		Botao Xiao,
		Muhammad Alrabeiah,
		Keyan Wang,
        Jun Chen
}

%
\author{Zheng Liu, Botao Xiao, Muhammad Alrabeiah, Keyan Wang, Jun Chen\thanks{ Zheng Liu, Botao Xiao, Muhammad Alrabeiah, and Jun Chen are with the Department of Electrical and Computer Engineering, McMaster University, Hamilton, ON L8S 4K1, Canada (email: \{liuz156, xiaob6, alrabm, chenjun\}@mcmaster.ca).} \thanks{Keyan Wang is with the School of Tele-communication Engineering, Xidian University, Xian, 710071, China (email: kywang@mail.xidian.edu.cn).}}
%

\markboth{Journal of \LaTeX\ Class Files,~Vol.~14, No.~8, August~2015}%
{Shell \MakeLowercase{\textit{et al.}}: Bare Demo of IEEEtran.cls for IEEE Journals}
%



\maketitle

\begin{abstract}
Haze and smog are among the most common environmental factors impacting image quality and, therefore, image analysis. This paper proposes an end-to-end generative method for image dehazing. It is based on designing a fully convolutional neural network to recognize haze structures in input images and restore clear, haze-free images. The proposed method is agnostic in the sense that it does not explore the atmosphere scattering model. Somewhat surprisingly, it achieves superior performance relative to all existing state-of-the-art methods for image dehazing even on SOTS outdoor images, which are synthesized using the atmosphere scattering model.
\end{abstract}

\begin{IEEEkeywords}
Convolutional neural network, image dehazing, image restoration,  residual learning.
\end{IEEEkeywords}

%
\IEEEpeerreviewmaketitle

\section{Introduction}

\IEEEPARstart{M}{any} modern applications rely on analyzing visual data to discover patterns and make decisions. Some examples could be found in intelligent surveillance, tracking, and control systems, where good quality images or frames are essential for accurate results and reliable performance. However, such systems could be significantly affected by environmentally induced distortions, the most common of which are haze and smog. Therefore, a lot of research in the computer vision community has been dedicated to addressing the problem of restoring good-quality images from their hazy counterparts, \cite{ColorAtten,DehazeNet,DarkChanPrior,berman2016non} to name a few. That problem is commonly referred to as the \textit{dehaze problem}.

The relation between the original and hazy images  \cite{narasimhan2002vision} is approximately captured by the following equation known as the atmosphere scattering model:
\begin{equation}\label{Physical}
I(x_{i})=J(x_{i})t(x)+A(1-t(x))\quad i=1,2,3 ,
\end{equation}
where for a pixel in the $i$th color channel and spatially indexed by $x$, $I(x_{i})$ is the intensity of the hazy pixel, $J(x_{i})$ is the actual intensity of that pixel, and $t(x)$ is the medium transmission function that depends on the scene depth and the atmospheric scattering coefficient $\beta$. Parameter $A$ in Equation (\ref{Physical}) is the atmosphere light intensity, which is assumed to be a global constant over the whole image. Since all variables in Equation (\ref{Physical}) are unknown except the hazy pixel intensity $I(x_{i})$, dehaze is in general an undetermined problem.

Over the past couple of decades, many methods have been proposed to solve the dehaze problem. Those methods could be loosely grouped into two categories: \textit{traditional} and \textit{Machine Learning (ML)-based} methods. The likes of \cite{DarkChanPrior}, \cite{ColorAtten}, and \cite{MarkovRandField} are some examples of the first category. They solve the underdetermined problem by exploiting some form of  prior information.

On the other hand, works such as \cite{RandForstReg}, \cite{DehazeNet}, \cite{MultiScaleCNN}, and \cite{AllInOne} have followed a learning-based approach. They leverage the advances in classic and deep learning technologies to tackle the dehaze problem. Regardless how different those two categories may seem, they all aim to recover the original image by first estimating the unknown parameters $A$ and $t(x)$ and then inverting Equation (\ref{Physical}) to determine $J(x_{i})$:
\begin{equation}
J(x_{i})=\frac{I(x_{i})-A(1-t(x))}{t(x)}\quad i=1,2,3.
\end{equation}

\begin{figure}
	\subfigure{
		\includegraphics[width=4cm]{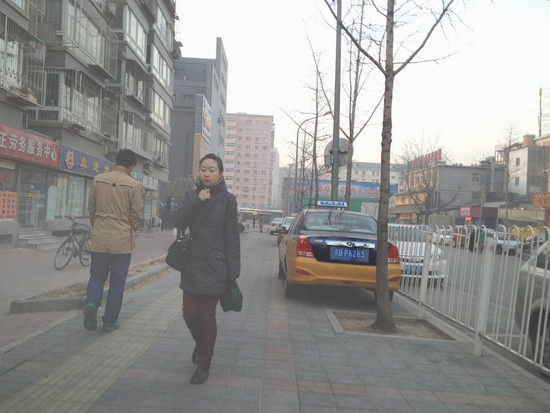}}
	\hspace{0in}
	\subfigure{
		\includegraphics[width=4cm]{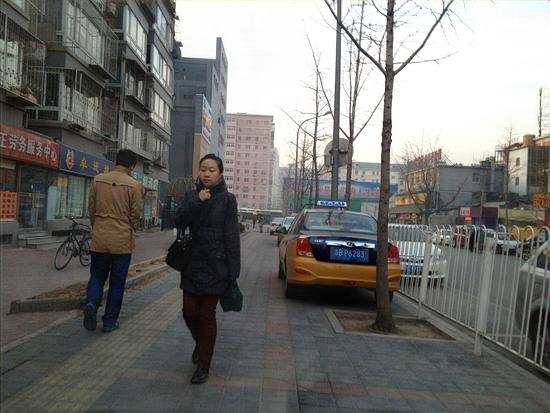}}
	\captionsetup{justification=centering}
	\centering\caption{Dehazing result of synthetic image. Left: Hazy input. Right: Clear output.}
\end{figure}

From the  viewpoint of estimation theory, the methods in both categories fall under the umbrella of the plug-in principle\footnote{Consider a parametric model $\mathcal{P}=\{P_{\theta}:\theta\in\Theta\}$ and a mapping $\tau:\Theta\rightarrow\mathbb{R}$. Suppose the observation comes from $P_{\theta^*}$. The plug-in principle refers to the method of constructing an estimate of $\tau(\theta^*)$ by first deriving an estimate of $\theta^*$, denoted by $\hat{\theta}$, then plugging $\hat{\theta}$ into $\tau(\cdot)$.}, and they will all be referred to as plug-in methods. 
However, for the dehaze problem, the optimality of the plug-in principle is not completely justified. Indeed, it is unlikely that the problem of lossy reconstruction of the original image can be transformed equivalently to an estimation problem for  parameters $A$ and $t(x)$ (or their variants), at least when the two problems are subject to the same evaluation metric. Moreover, the actual relation between the original and hazy images  can be fairly complex and may not be fully captured by the atmosphere scattering model. Due to this potential mismatch, methods that rely on the atmosphere scattering model (including but not limited to plug-in methods) do not guarantee desirable generalization to natural images even if they can achieve good performance on synthetic images.







Based on the aforementioned take on plug-in methods (and, more generally, model-dependent methods), this paper approaches the dehaze problem from a different, and more \textit{agnostic}, angle; it presents a dehaze neural network that solely focuses on producing a haze-free version of the input image. It utilizes the recent advances in deep learning to build an encoder-decoder network architecture that is  trained to directly restore the clear image, ignoring the parameter estimation problem altogether. The proposed method also has the potential of recognizing complex haze structures present in the training data but not captured by the atmosphere scattering model. To the best of our knowledge, such view of the dehaze problem has never been explored except in the recent work  \cite{GFN}, where a so-called Gated Fusion Network (GFN) is introduced for image dehazing. It will be seen that our proposed network has several advantages over GFN, especially in terms of architecture complexity and input-size flexibility; moreover, certain characteristics of GFN are specifically tailored to the dehaze problem whereas the architecture of our network is more generic and consequently more broadly applicable. 

The rest of the paper is organized into three sections. The following section, Section 2, presents a Generic Model-Agnostic convolutional neural Network (GMAN) for image dehazing together with a detailed explanation of  the network architecture and its building blocks. Section 3 will introduce the experimental results, showing the performance of the proposed GMAN. It also includes a description of the dataset and training procedure. Finally, Section 4 will wrap up the paper with some concluding remarks.

%
%
%
%



%

\begin{figure*}
	\centering
	\includegraphics[width=0.8\linewidth]{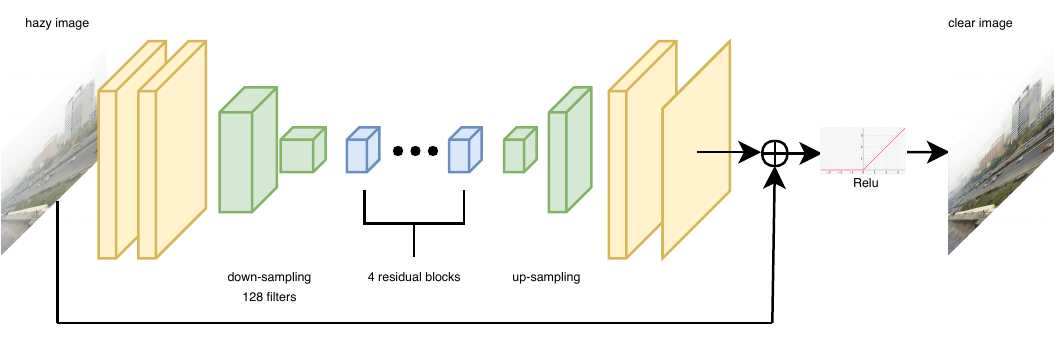}
	\captionsetup{justification=centering}
	\centering\caption{Structure and details of GMAN. The yellow blocks are convolutional layers, the green blocks are down-sampling layers and deconvolutional layers. We cascade 4 residual blocks shown as blue blocks, and the number of convolutional layers inside are 2, 2, 3, 4.}
	\label{network structure}
\end{figure*}

\section{Proposed Method}

Since the single image haze removal is an ill-posed problem, a deep neural network based on convolutional, residual, and deconvolutional blocks is devised and trained to take on a hazy image and restore its haze-free version. The network has an encoder-decoder structure as shown in Fig. \ref{network structure}. In the following subsections, the network architecture, its building blocks, and the training loss function are discussed in more detail.

\subsection{Network Architecture}

\begin{figure}
	\centering
 \includegraphics[width=0.252\textwidth,height=0.196\textheight]{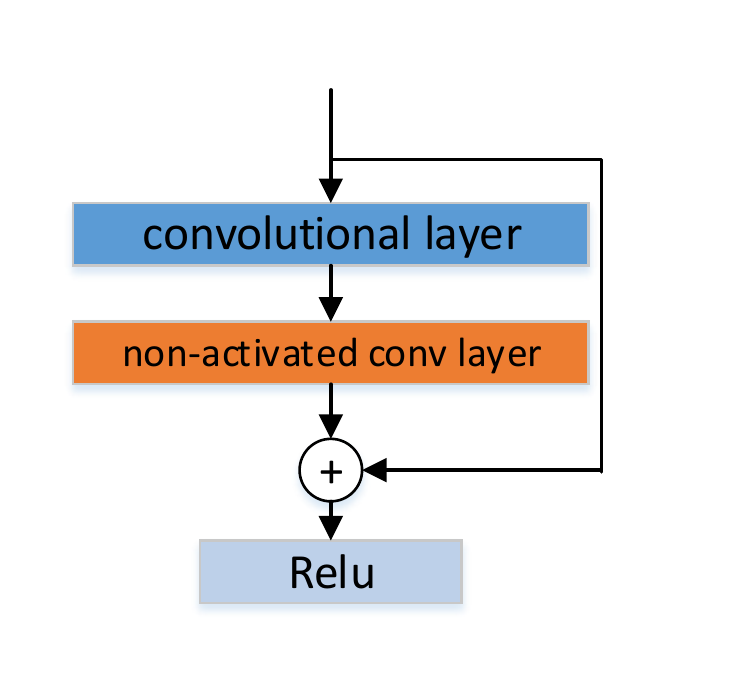}
	\captionsetup{justification=centering}
	\centering\caption{A residual block used in the middle layer of the proposed GMAN. In each block, the number of convolutional layers can be different. Relu is used as the activation function after the addition operator of every block.}
	\label{resblock}
\end{figure}

The proposed network is a fully Convolutional Neural Network (CNN). It is used to restore a clear image from a hazy input one. Functionally speaking, it is an end-to-end generative network that uses encoder-decoder structure with down- and up-sampling factor of 2. Its first two layers are constructed with 64-channel convolutional blocks. Following them are two-step down-sampling layers that encode the input image into a $56\times56\times 128$ volume. The encoded image is then fed to a residual layer built with 4 residual blocks, each containing a shortcut connection, see Fig. \ref{resblock}. This layer represents the transition from encoding to decoding, for it is followed by the deconvolutional layer that up-samples the residual layer output and reconstructs a new $224\times224\times64$ volume for another round of convolutions. The last two layers comprise convolutional blocks. They transform the up-sampled feature maps into an RGB image, which is finally added to the input image and thresholded with a ReLU to produce the haze-free version.

\subsection{Residual Learning}


The network uses residual learning on two levels, local and global. In the middle layer and just right after down-sampling, the residual blocks are used to build the local residual layer. It takes advantage of the hypothesized and empirically proven \cite{kim2016accurate,zhang2017beyond,szegedy2017inception,ren2017faster} easy-to-train property of residual blocks (see \cite{ResNet}), and learns to recognize haze structures. Residual learning also appears in the overall architecture of the proposed GMAN. Specifically, the input image is fed along with the output of the final convolutional layer to a sum operator, creating one global residual block, see Fig. \ref{network structure}. The main advantage of this global residual block is that it helps the proposed network better capture the boundary details of objects with different depths in the scene.

\subsection{Encoder-Decoder Architecture}

The architecture of the proposed GMAN follows the popular encoder-decoder architecture used in the deniosing problem. It is composed of three parts: encoder, hidden layers, and decoder. This architecture makes it possible to train a deep network and decrease the dimension of data. Since haze could be thought of as a form of noise, the encoder output is down-sampled and fed to the residual layer to extract important features. The network squeezes out the features of the original image and discards of the noise information. The decoder part is expected to learn and regenerate the missing data of the haze-free image, conforming the statistical distribution of the input information during the decoding period.

\begin{figure*}
	\includegraphics[width=0.13\linewidth]{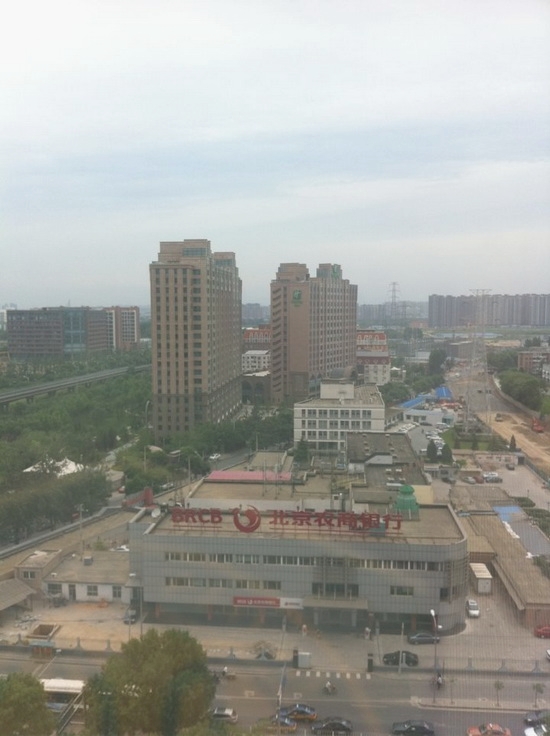}
	\includegraphics[width=0.13\linewidth]{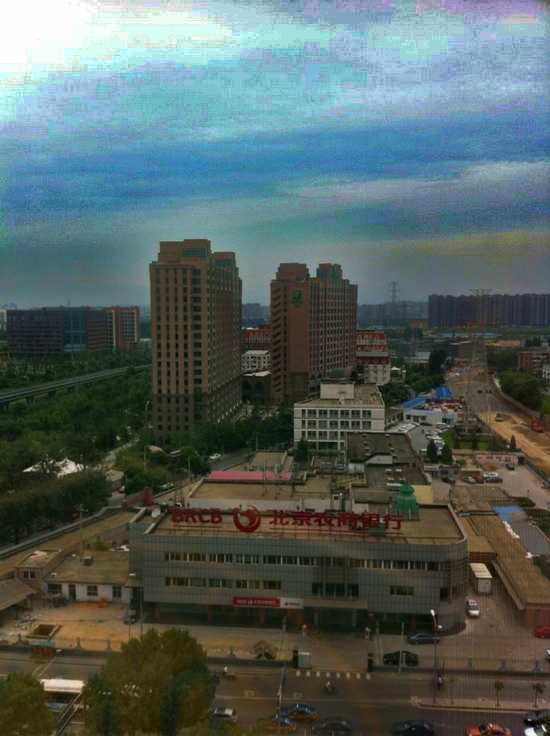}
	\includegraphics[width=0.13\linewidth]{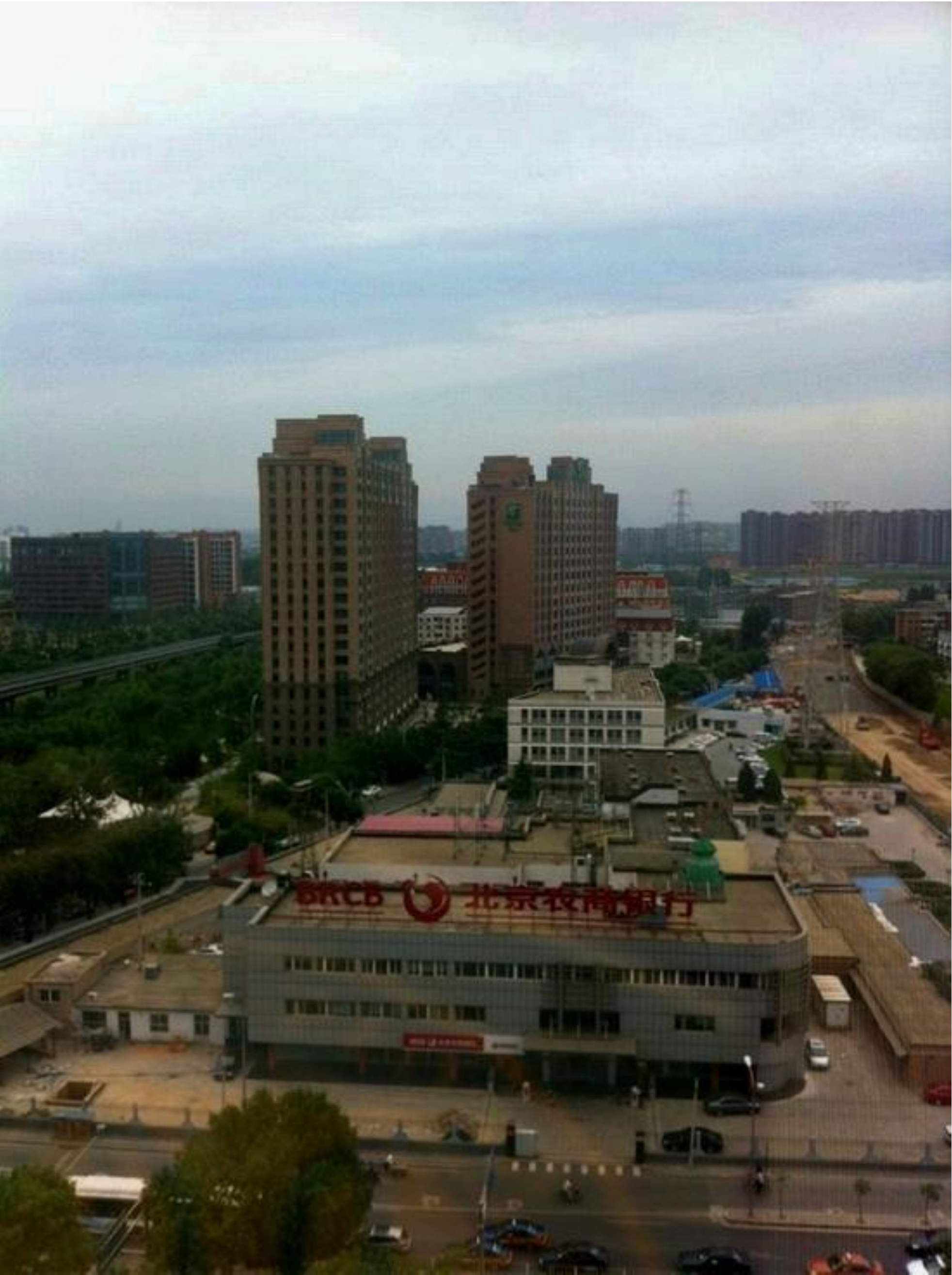}
	\includegraphics[width=0.13\linewidth]{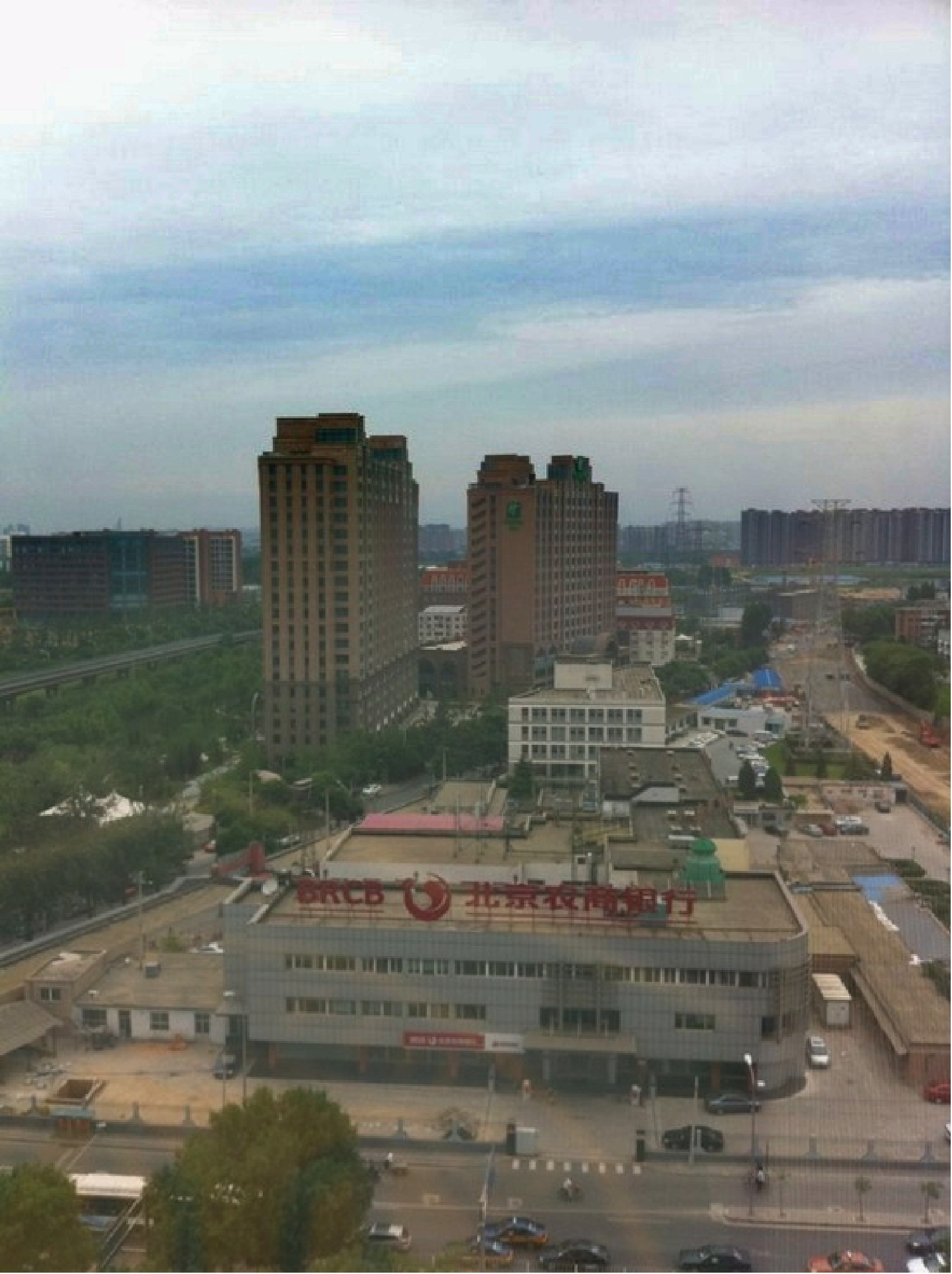}
	\includegraphics[width=0.13\linewidth]{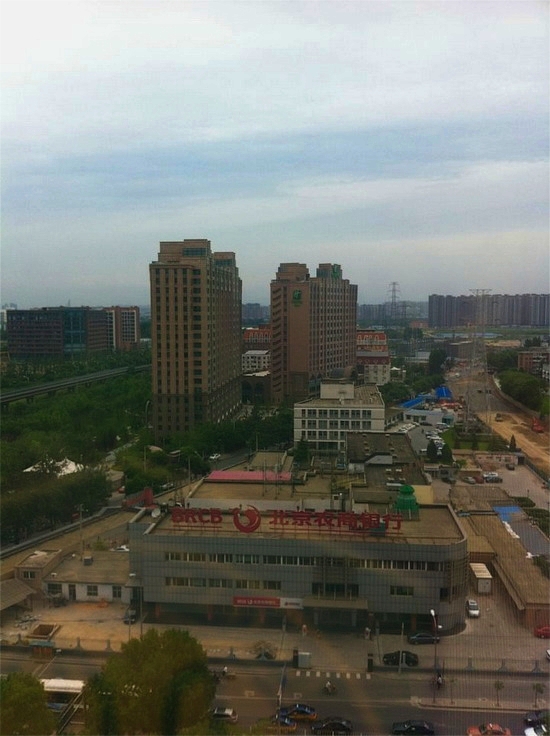}
	\includegraphics[width=0.13\linewidth]{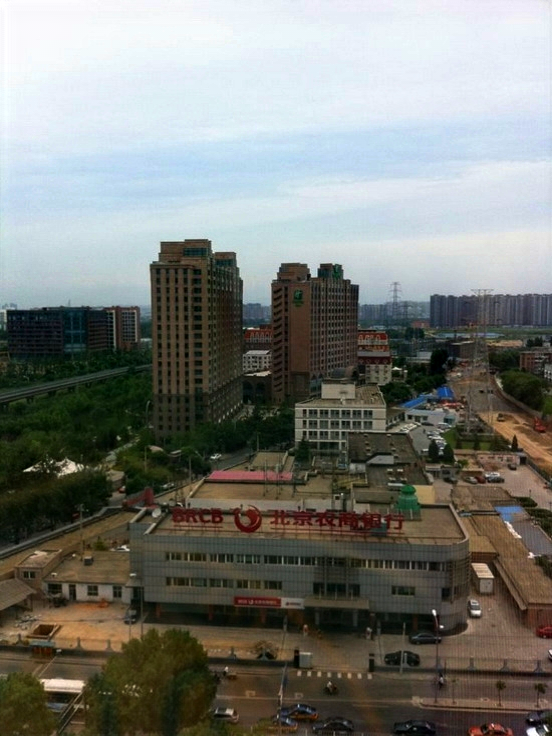}
	\includegraphics[width=0.13\linewidth]{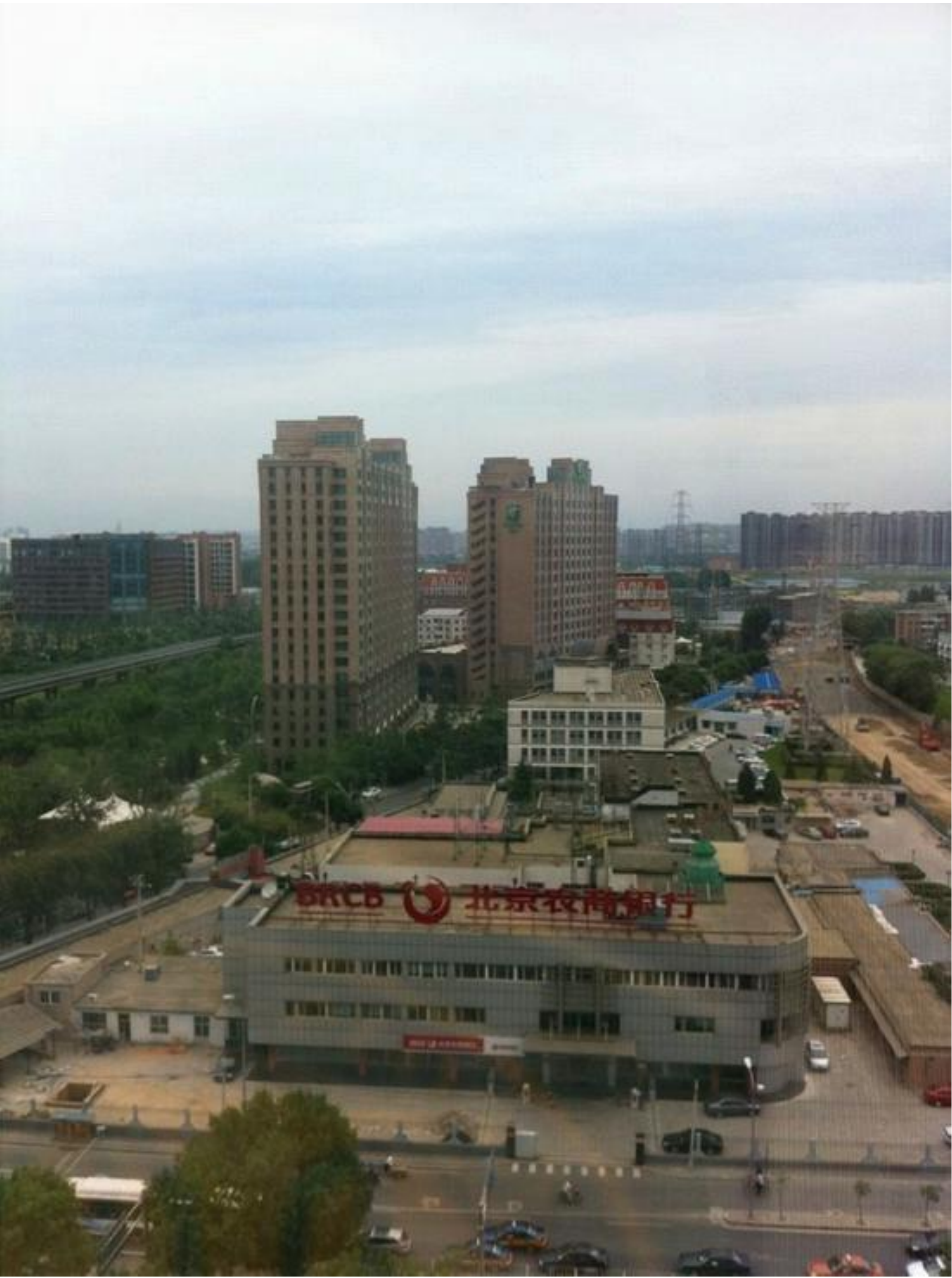}
	\centering
\end{figure*}
\begin{figure*}
	\subfigure[Hazy]{\includegraphics[width=0.13\linewidth]{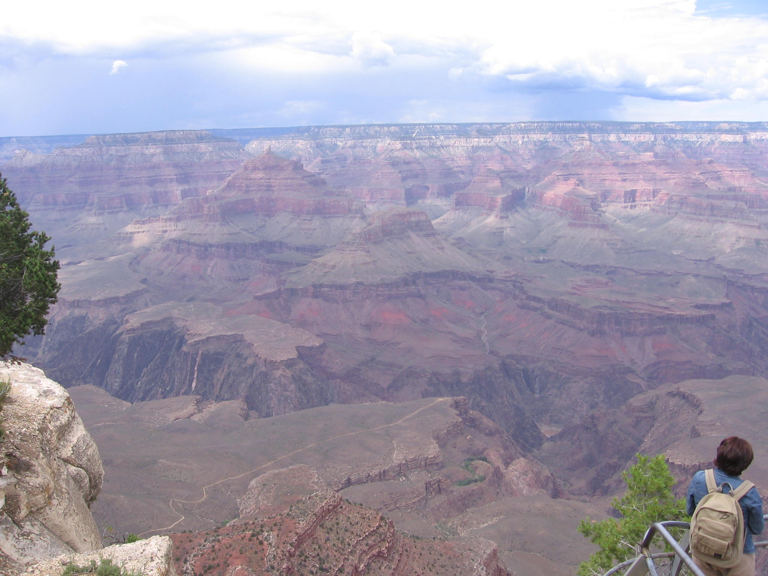}}
	\subfigure[DCP]{\includegraphics[width=0.13\linewidth]{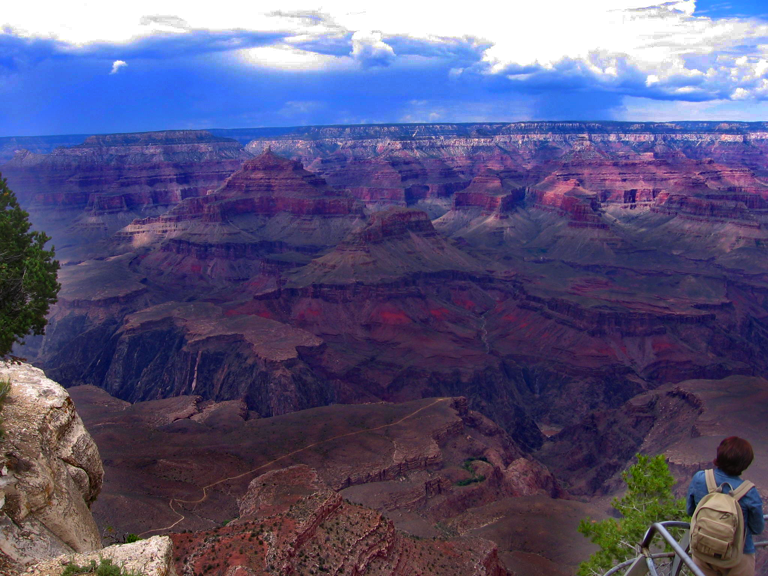}}
	\subfigure[Dehazenet]{\includegraphics[width=0.13\linewidth]{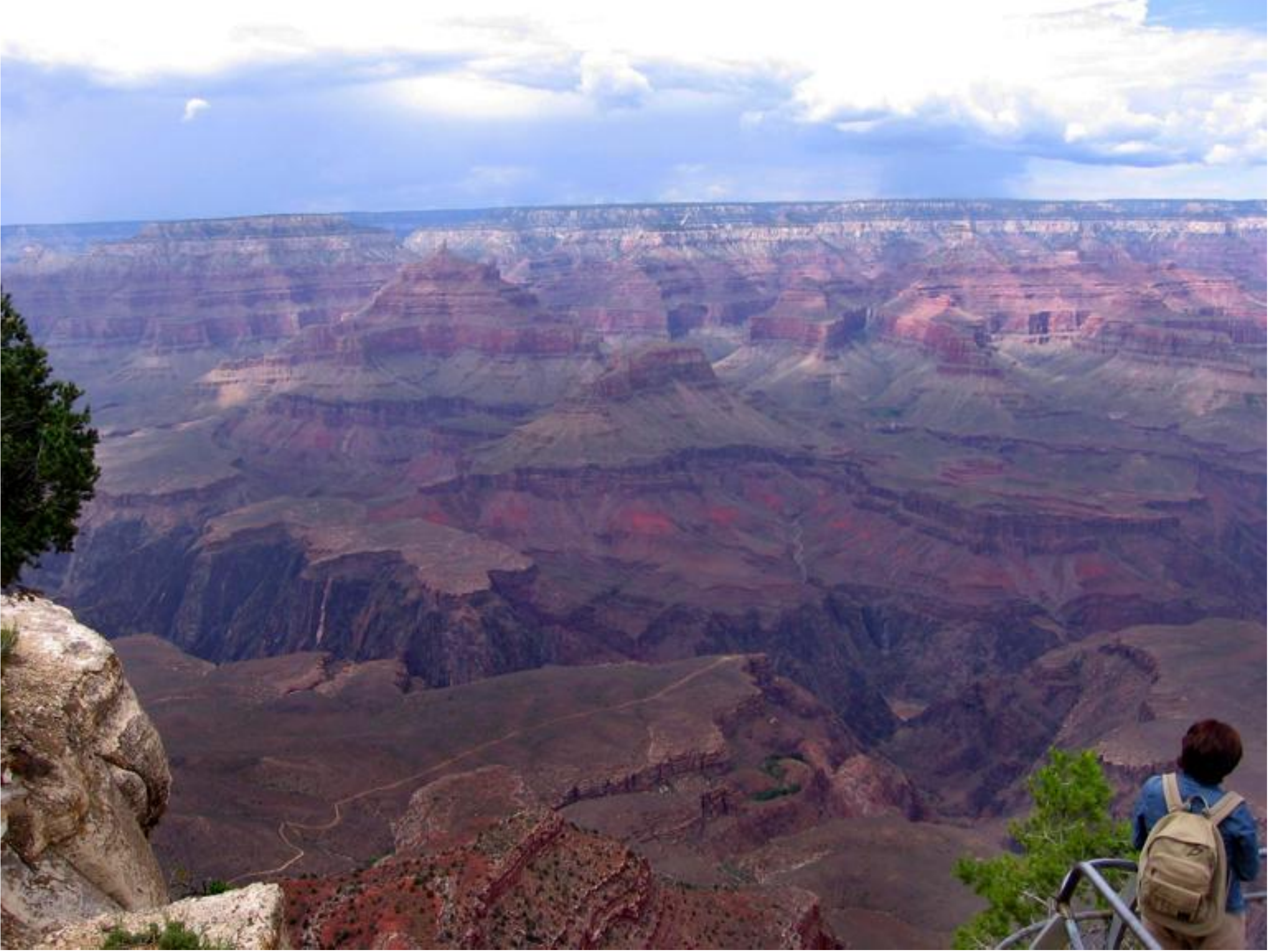}}
	\subfigure[MSCNN]{\includegraphics[width=0.13\linewidth]{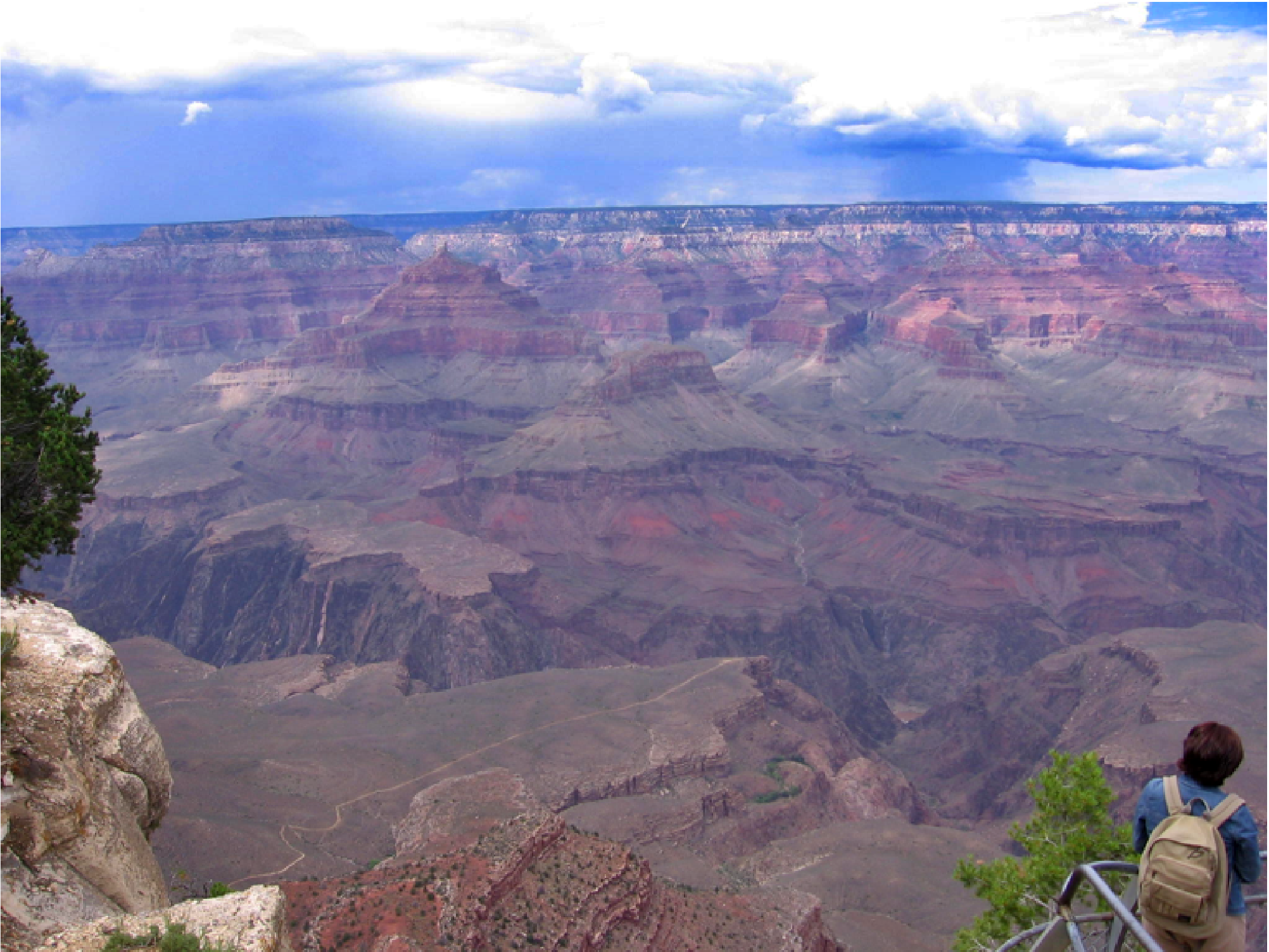}}
	\subfigure[AOD-Net]{\includegraphics[width=0.13\linewidth]{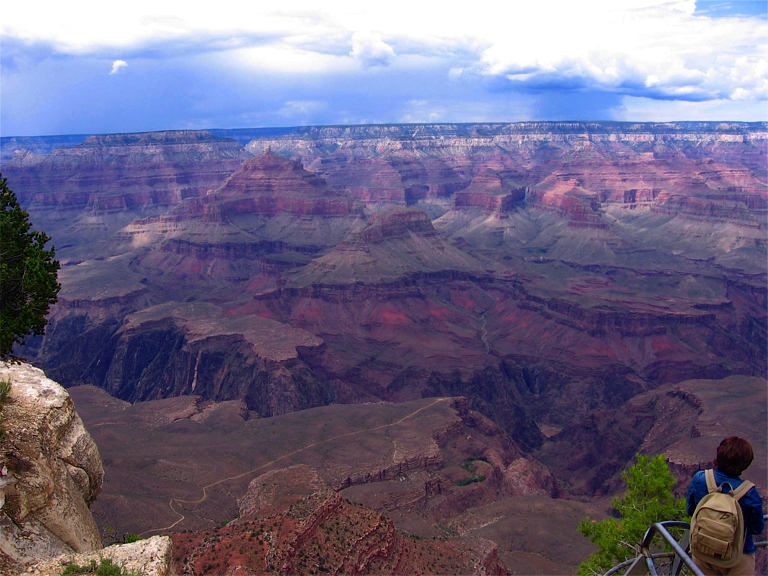}}
	\subfigure[GFN]{\includegraphics[width=0.13\linewidth]{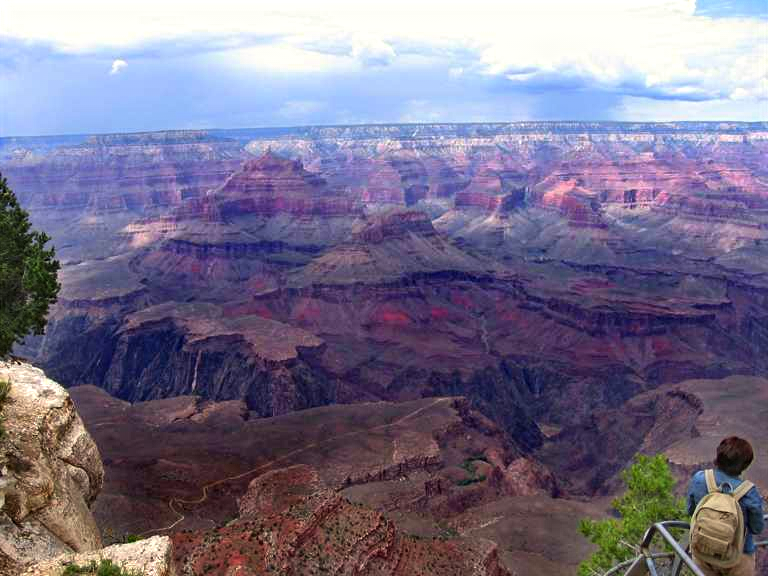}}
	\subfigure[GMAN]{\includegraphics[width=0.13\linewidth]{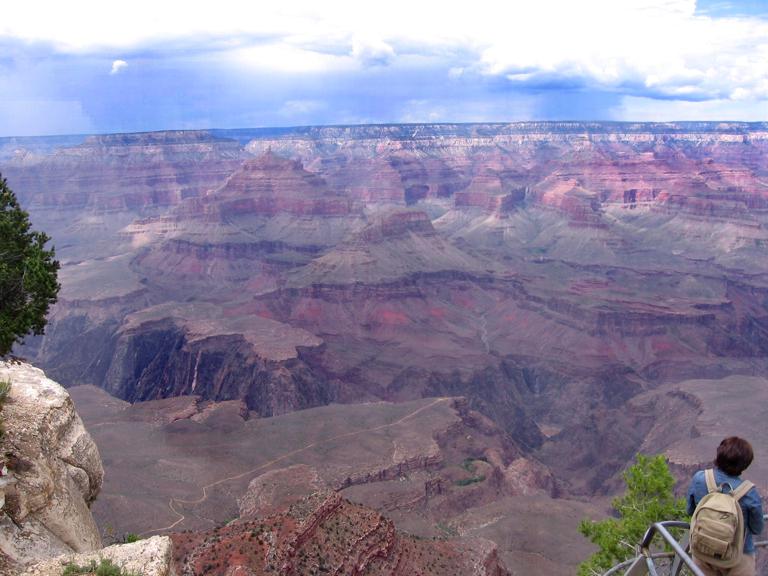}}
	\captionsetup{justification=centering}
	\centering
	\caption{Comparison of different dehaze methods. First row has examples of synthetic hazy images. Second row has examples of natural hazy images.}
	\label{figComp}
\end{figure*}

\subsection{Loss Function: MSE and Perceptual Loss}

To train the proposed GMAN, a two-component loss function is defined. The first component measures the similarity between the output and the ground truth, and the second helps produce a visually meaningful image. The following three subsections provide more information on each component and the total loss:

\subsubsection{MSE Loss}

Using PSNR to measure the difference between the output image and the ground truth is the most common way to show the effectiveness of an algorithm. Thus, MSE is chosen to be the first component of the loss function, namely $L_{MSE}$. The optimal value of PSNR could be reached by minimizing MSE at pixel level, which is expressed as:
\begin{equation}
L_{MSE}=\frac{1}{N}\sum_{x=1}^{N}\sum_{i=1}^{3}\parallel\hat{J}(x_{i})-J(x_{i})\parallel^2,
\end{equation}
where $\hat{J}(x_{i})$ is the output of the network, $J(x_{i})$ is the ground truth, $i$ is the channel index, and $N$ is the total number of pixels.

\subsubsection{Perceptual Loss}

In many classic image restoration problems, the quality of the output image is measured solely by the MSE loss. However, the MSE loss is not necessarily a good indicator of the visual effect. As Johnson \textit{et al.} demonstrate in \cite{johnson2016perceptual}, extracting high level features from specific layers of a pre-trained neural network can be of benefit to content reconstruction. The perceptual loss obtained from high-level features can describe the difference between two images more robustly than pixel-level losses.

Adding a perceptual loss component enables the decoder part of GMAN to acquire an improved ability to generate fine details of target images using features that have been extracted. In the present work, the network output and the ground truth are both fed to VGG16 \cite{VGG16}; following \cite{johnson2016perceptual}, we use the feature maps extracted from layers \textit{$conv1_1, conv2_2, conv3_3$} (which will be simply referred to as layers 1, 2, 3) of VGG16 to define the perceptual loss $L_{p}$ as follows:
\begin{equation}\label{perceptual loss}
L_{p}=\sum\limits_{j=1}^3\frac{1}{C_{j}H_{j}W_{j}}\parallel\phi_{j}(\hat{J})-\phi_{j}(J)\parallel^{2}_{2},
\end{equation}
where $\phi_j(\hat J)$ and $\phi_j(J)$ are the feature maps of  layer $j$ of VGG16 induced by the network output and the ground truth, respectively, and $C_j$, $H_j$, and $W_j$ are the dimensions of the feature volume of layer $j$ of VGG16.
 


\subsubsection{Total Loss}
Combining both MSE and perceptual loss components results in the total loss of GMAN. In order to provide some sort of balance between the two components, the perceptual loss is pre-multiplied with $\lambda$, yielding the following expression:
\begin{equation}\label{loss}
L=L_{MSE}+\lambda_{}L_{p}.
\end{equation}


\section{Experimental Result}

This section first describes the training dataset and procedure. It then presents an evaluation of the performance of the proposed GMAN\footnote{The relevant codes can be found at https://github.com/Seanforfun/Deep-Learning/tree/master/DehazeNet.}, which is being benchmarked to some of the existing methods.

\subsection{Dataset}

According to the atmosphere scattering model, the transmission map $t(x)$ and atmosphere light intensity $A$ control the haze level of an image. Therefore, setting these two factors properly is important for building a dataset of hazy images. We use the OTS dataset from RESIDE \cite{li2017reside}, which is built using collected real-world outdoor scenes. The whole dataset contains 313,950 synthetic hazy images, generated from 8970 ground-truth images by varying the values of $A$ and $\beta$ (the depth information is estimated using \cite{liu2016learning}).
  Thus, for each ground-truth image, there are 35 corresponding hazy images. We notice that the testing set of RESIDE, the SOTS, has 1000 ground-truth images, each with 35 synthetic hazy counterparts, that are all contained in the training data. This certainly can lead to some inaccuracies in testing results. Thus, the testing images were all removed from the training data (including their hazy counterparts), leading to a reduced-size training dataset of 278,950 hazy images (generated from 7970 ground-truth images).

\subsection{Training}

The proposed GMAN is trained end-to-end by minimizing the loss $L$ given by Equation (\ref{loss}). All layers in GMAN have 64 filters (kernels), except for the down-sampling ones which have 128 filters, with spatial size of $3\times3$. The network requires an input with size $224\times224$, so every image in the training dataset is randomly cropped in order to fit the input size\footnote{This restriction is only for the training phase. The trained network can be applied to images of arbitrary size}. The batch size is set to 35 to balance the training speed and the memory consumption on the GPU. For accelerated training, the Adam optimizer \cite{AdamOpt} is used with the following settings: the initial learning rate of 0.001, $\beta_1=0.9$, and $\beta_2=0.999$. The network and its training process have been implemented using TensorFlow software framework and carried out on an NVIDIA Titan Xp GPU. After 20 epochs of training, the loss function drops to a value of 0.0004, which is considered a good stopping point.

\subsection{Evaluation Results}

The proposed GMAN achieves superior performance relative to many state-of-the-art methods. According to Table \ref{t.1} \footnote{In Tables \ref{t.1} and \ref{t.2}, the performance results of other methods except GFN are quoted from \cite{li2017reside}.} below, it clearly outperforms all other competing methods under consideration on the SOTS outdoor dataset \cite{DarkChanPrior,DehazeNet,MultiScaleCNN,AllInOne}.  Moreover, as shown in Fig. \ref{figComp}, GMAN avoids darkening the image color  as well as the excessive sharpening of object edges. In contrast, it can be seen from Fig. \ref{figComp} that the DCP  method \cite{DarkChanPrior} dims the light intensity of the dehazed image, and causes color distortions in high-depth-value regions (e.g., sky); though MSCNN \cite{MultiScaleCNN} does well in these high-depth-value regions, its performance degrades in medium-depth areas of the target image. Hence, the proposed GMAN can overcome many of these issues and generate a better haze-free image.

We have also tested our network on the SOTS indoor dataset (see Table \ref{t.2}). In this case, the performance is not as impressive, and comes fourth after DehazeNet, GFN, and AOD-Net. Nevertheless, one can still see the great promise of the model-agnostic dehaze methods even on the indoor dataset. Indeed, also as a member of the family of model-agnostic networks, GFN is ranked second in terms of PSNR and ranked first (almost tied with the top-ranked DehazeNet) in terms of SSIM. Our preliminary results indicate that it is possible to design a more powerful model-agnostic network  that dominates all the existing ones (especially those based on the plug-in principle) on both SOTS outdoor and indoor datasets by integrating and generalizing the ideas underlying GMAN and GFN. This line of research will be reported in a followup work.

\begin{table}[!htbp]
	\centering
	\caption{Performance comparison on the SOTS outdoor dataset.}
	\begin{tabular}{ccccccc}
		\hline
		&DCP &DehazeNet &MSCNN &AOD-Net &GFN &GMAN\\ \hline
		PSNR &18.54 &26.84 &21.73 &24.08 &21.67 &28.19\\
		SSIM &0.7100 &0.8264 &0.8313 &0.8726 &0.8524 &0.9638\\ \hline
	\end{tabular}
	\label{t.1}
\end{table}

\begin{table}[!htbp]
	\centering
	\caption{Performance comparison on the SOTS indoor dataset.}
	\begin{tabular}{ccccccc}
	\hline
		&DCP &DehazeNet &MSCNN &AOD-Net &GFN &GMAN\\ \hline
		PSNR &18.87 &22.66 &20.01 &21.01 &22.44 &20.53\\
		SSIM &0.7935 &0.8325 &0.7907 &0.8372 &0.8844 &0.8081\\ \hline
	\end{tabular}
	\label{t.2}
\end{table}

\section{Conclusion}
The proposed GMAN in this paper explores a new direction of solving the dehaze problem. With its encoder-decoder fully convolutional architecture, GMAN learns to capture haze structures in images and restore the clear ones without referring to the atmosphere scattering model. It also avoids the deemed-unnecessary estimation of parameters $A$ and $t(x)$. Experimental results have verified the potential of GMAN in generating haze-free images and shown that it is capable of overcoming some of the common pitfalls of state-of-the-art  methods, like color darkening and excessive edge sharpening. Moreover, due to the generic architecture of GMAN, it could lay the groundwork for further research on general-purposed image restoration. Indeed, we expect that through training and some design tweaks, our network could be generalized to capture various types of image noise and distortions. In this sense, the present work not only suggests an improved solution to the dehaze problem, but also represents a progressive move towards developing a universal image restoration method.


%

\appendices


%
%

\ifCLASSOPTIONcaptionsoff
  \newpage
\fi



\bibliographystyle{IEEEtran}
%

\bibliography{IEEEabrv,citation}


%








\end{document}